\documentclass[runningheads]{llncs}

\usepackage{eccv}

\usepackage{enumitem}

\usepackage{multirow}

\usepackage{pifont}
\newcommand{\cmark}{\ding{51}}%
\usepackage{eccvabbrv}
\usepackage{arydshln}
\usepackage{xcolor} 
\definecolor{deepgreen}{RGB}{0,100,0} 
\definecolor{darkblue}{RGB}{0,0,180} 

\usepackage{graphicx}
\usepackage{booktabs}

\usepackage[accsupp]{axessibility}  

\usepackage[pagebackref,breaklinks,colorlinks,citecolor=eccvblue]{hyperref}
\usepackage{hyperref}

\usepackage{orcidlink}
\newcommand{\ours}{\textbf{\textsf{Co-Instruct}}}
\newcommand{\mic}{\textbf{\textsf{MICBench}}}
\newcommand{\gb}[1]{\textbf{\textcolor{deepgreen}{#1}}}

\begin{document}

\title{Towards Open-ended Visual Quality Comparison} 


\titlerunning{Towards Open-ended Visual Quality Comparison}

\author{Haoning Wu\thanks{Equal contribution. $^\text{1}$Nanyang Technological University. $^\text{2}$City University of Hong Kong. $^\text{3}$Shanghai Jiao Tong University. $^\text{4}$Sensetime Research.}\inst{1} \and
Hanwei Zhu$^\star$\inst{2} \and
Zicheng Zhang$^\star$\inst{3} \and Erli Zhang\inst{1} \\ Chaofeng Chen\inst{1} \and Liang Liao\inst{1} \and Chunyi Li\inst{3} \and Annan Wang\inst{1}, Wenxiu Sun\inst{4} \\ Qiong Yan\inst{4} \and Xiaohong Liu\inst{3} \and Guangtao Zhai\inst{3} \and Shiqi Wang\inst{2} \and Weisi Lin\inst{1} 
}
\authorrunning{H. Wu et al.}

\institute{
\url{https://huggingface.co/q-future/co-instruct}}

\maketitle

\begin{figure}
    \vspace{-2.8em}

    \centering
    \includegraphics[width=\linewidth]{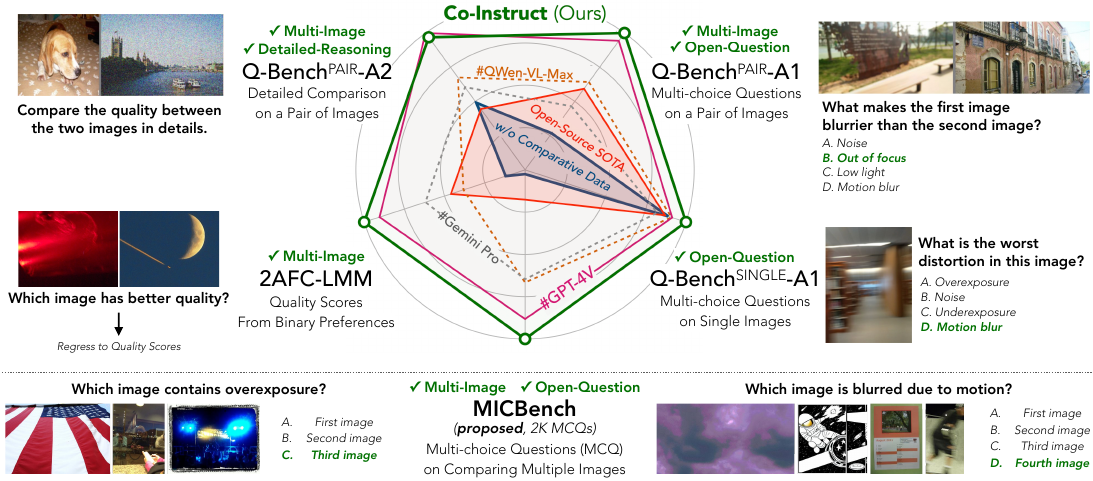}
    \vspace{-2.0em}
    \caption{The proposed \ours, \textit{first-of-its-kind} open-source LMM with capability on \textcolor{deepgreen}{\textit{open-question}}~\&~\textcolor{deepgreen}{\textit{detailed-reasoning}} visual quality comparison. It outperforms existing LMMs on the proposed \textbf{\textsf{MICBench}} as well as existing quality evaluation benchmarks.}
    \label{fig:radar}
    \vspace{-2.5em}
\end{figure}


\begin{abstract}

Comparative settings (\eg \textit{pairwise choice, listwise ranking}) have been adopted by a wide range of subjective studies for image quality assessment (IQA), as it inherently standardizes the evaluation criteria across different observers and offer more clear-cut responses. In this work, we extend the edge of emerging large multi-modality models (LMMs) to further advance visual quality comparison into open-ended settings, that \textbf{1)} can respond to \textcolor{deepgreen}{\textit{open-range questions}} on quality comparison; \textbf{2)} can provide \textcolor{deepgreen}{\textit{detailed reasonings}} beyond direct answers. To this end, we propose the \ours. To train this \textit{first-of-its-kind} open-source open-ended  visual quality comparer, we collect the {Co-Instruct-562K} dataset, from two sources: \textbf{(a)} LLM-merged single image quality description, \textbf{(b)} GPT-4V \textit{``teacher''} responses on unlabeled data. Furthermore, to better evaluate this setting, we propose the \mic, the first benchmark on \underline{m}ulti-\underline{i}mage \underline{c}omparison for LMMs. We demonstrate that \ours~not only achieves in average \textbf{30\%} higher accuracy than state-of-the-art open-source LMMs, but also outperforms GPT-4V (\textit{its teacher}), on both existing related benchmarks and the proposed \mic. 

\end{abstract}

\begin{figure}

    \centering
    \includegraphics[width=\linewidth]{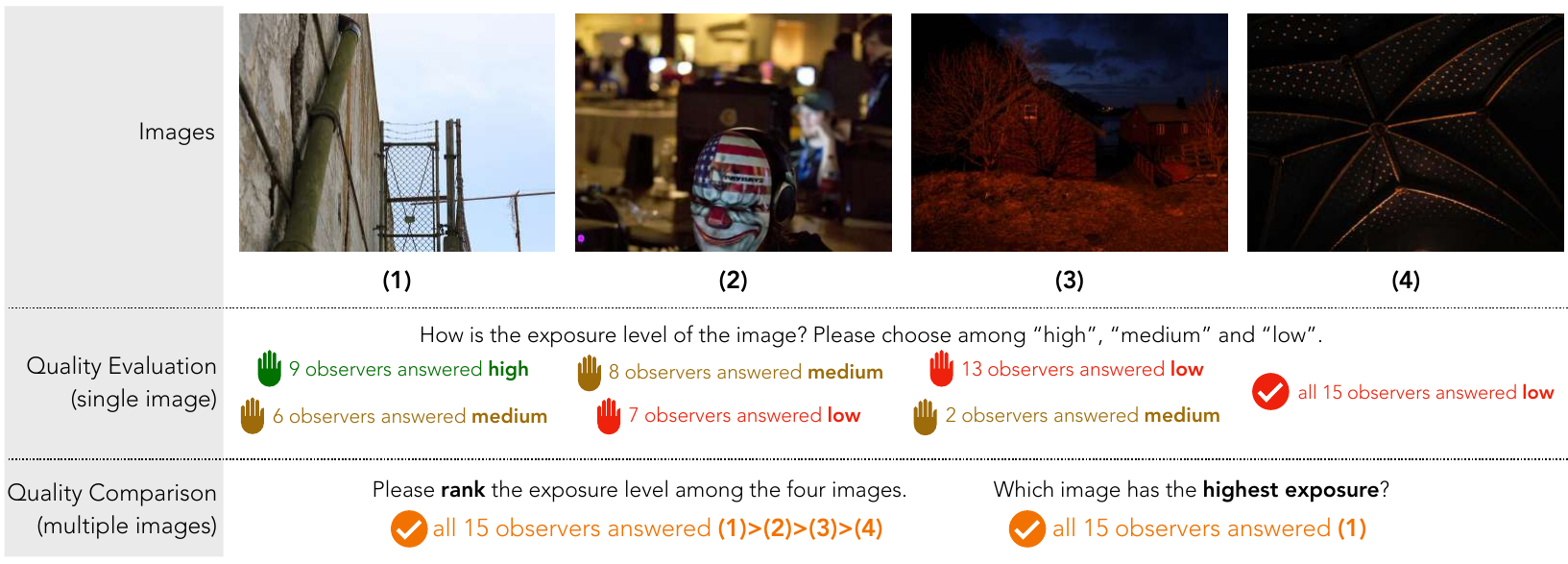}
    \vspace{-2.0em}
    \caption{\textbf{The motivation of} open-ended visual quality comparison: comparative settings can effectively avoid the \textbf{ambiguity on absolute evaluations} for single images, and provide more clear-cut judgements to serve as downstream guidances~\cite{imagereward,ranksrgan}.}
    \label{fig:problemdefinition}
    \vspace{-2.2em}
\end{figure}

\section{Introduction}
\label{sec:intro}

Image quality assessment (IQA) has been an important domain in visual computing, as it provides effective recommendation~\cite{ytugc} on high-quality visual contents and valuable guidance~\cite{imagereward,irpotential,ranksrgan} for potential improvements. Most recently, several pioneer studies~\cite{q-instruct,wu2024qbench,qalign} have explored large multi-modality models (LMMs, \eg GPT-4V)~\cite{openai2023gpt4,geminipro,improvedllava,mplug2, iblip, internlmxcomposer2}, on expanding IQA from giving a scalar score (\textit{e.g.} 3.457) to the {{open-ended}} scenarios, that allows evaluations in response to \textcolor{deepgreen}{\textit{open-range questions}}, and provide \textcolor{deepgreen}{\textit{detailed reasonings}} beyond an overall score. 

While these studies sufficiently \textit{emulate human ability} on IQA, they also suffer from the same drawback as human: \textbf{ambiguity on absolute evaluations}. For instance, as shown in Fig.~\ref{fig:problemdefinition}~(a), different human observers hold different standards on the exposure level on single images, and henceforth provide diversified absolute evaluations. Nevertheless, while asked to compare the exposure level of the images, all observers agree with the rank (1)>(2)>(3)>(4) (Fig.~\ref{fig:problemdefinition}(b)); all observers also agree that (1) has the highest exposure, though not all choose the option \textit{high} while evaluating it independently. Given this observation, comparison has been a traditional human study setting for quality evaluation~\cite{howtocompare} and adopted by a wide range of existing subjective studies~\cite{pieapp,lpips,pipal,imagereward,hps}. Furthermore, to avoid the ambiguity, the comparative settings are also predomintantly adopted~\cite{ranksrgan,imagereward} while applying IQA algorithms for improvement guidance.

\begin{figure*}[h]

    \centering
    \includegraphics[width=\linewidth]{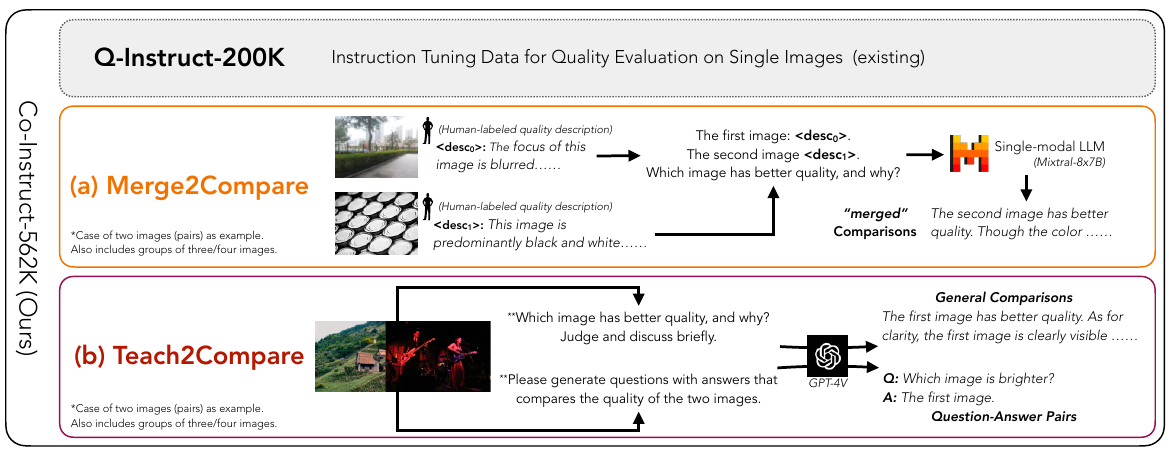}
    \vspace{-2.0em}
    \caption{\textbf{The construction methodology of} Co-Instruct-562K, a combination of \textbf{(a) Merge2Compare} (\textit{LLM comparison from human-labeled single image quality descriptions}) and \textbf{(b) Teach2Compare} (\textit{GPT-4V comparison on multiple unlabeled images}).}
    \label{fig:coinstructdbprinciple}
    \vspace{-2.2em}
\end{figure*}

While comparison has widely-recognized significance for IQA, existing related datasets~\cite{pieapp} and methods~\cite{zhang2021uncertainty,rankiqa} are generally based on overall quality comparison and have not extended to the open-ended scenarios; on the other hand, open-source LMMs~\cite{q-instruct,improvedllava,iblip} are usually only fine-tuned with \textit{single image} instruction tuning datasets~\cite{llava,gqa,cocovqa} and proved to lack enough capability even on two image comparison settings~\cite{qbenchplus,2afc}. While these gaps have clearly indicated the need of a specific instruction tuning dataset for visual quality comparison, it is too expensive to collect such a dataset from human. To avoid costly human labors, we propose an alternative strategy to collect the training dataset, named \textit{\underline{Co}llaborative \underline{Instruct}ion Tuning from Weak Supervisors} (\textit{Co-Instruct}). Specifically, we adopt two non-perfect supervisors: \textbf{1)} \textbf{Merge2Compare} (Fig.~\ref{fig:coinstructdbprinciple}(a)). Originated from single image quality descriptions on 19K images as labeled by human in the Q-Pathway~\cite{q-instruct} dataset, we randomly match them into 100K groups (\textit{2-4 images per group}) with removing the most similar descriptions with an text embedding model~\cite{e5mistral}. Then, we prompt a single-modal large language model (LLM)~\cite{mixtral} to compare multiple human descriptions in a group, and ``\textit{merge}'' them into 100K pseudo comparisons. \textbf{2)} \textbf{Teach2Compare} (Fig.~\ref{fig:coinstructdbprinciple}(b)). Observing that GPT-4V has especially high accuracy on pairwise settings~\cite{2afc,qbenchplus} among existing LMMs, following existing practices~\cite{sharegpt4v,laion2023gpt4v}, we leverage GPT-4V responses to further expand our dataset. We collect 9K unlabeled images and randomly match into 30K image groups (\textit{also 2-4 images per group}), and obtain GPT-4V responses on both \textit{caption-like} general comparisons and \textit{question-answer pairs} for comparisons. 
By integrating Q-Instruct-200K~\cite{q-instruct} (\textit{on single images}), \textbf{Merge2Compare}, and \textbf{Teach2Compare}, we construct the Co-Instruct-562K, the first instruction tuning dataset for open-ended visual quality comparison.

To correctly refer to each specific image during conversation, we define a specific image-text interleaved format~\cite{internlmxcomposer2} to handle multi-image cases, as follows: \\
{{\tt~~User: The first image: \textcolor{darkblue}{<img$_\texttt{0}$>} The second image: \textcolor{darkblue}{<img$_\texttt{1}$>} ... <query> \\}}{{\tt~~Assistant: <response>}}

Moreover, as we need to feed multiple images together to the LLM decoder, adopting the most popular LLaVA~\cite{llava,bakllava,improvedllava} structure that linearly projects visual embeddings will exceed the context window of the language models~\cite{llama2} and cause errors. Henceforth, we adopt an alternative visual abstractor structure~\cite{mplug2} to first reduce visual token length (\textit{from 1,025 to 65 tokens per image}), and then concatenate them with text embeddings to pass to language decoders. By learning from the Co-Instruct-562K dataset and the specially-designed input structure, we present the \ours, with up to \textbf{86\%} improvements than its baseline~\cite{mplug2}, and \textbf{61\%} better than the state-of-the-art open-source LMM. 
More importantly, despite using GPT-4V as one of its teachers, it still surpasses the GPT-4V \textit{teacher} in a variety of multi-choice question (MCQ) benchmarks, and matches GPT-4V ability in scenarios requiring detailed language reasonings.

After training the model \ours, our another concern is the lack of abundant evaluation settings on multi-image comparison: while Q-Bench~\cite{wu2024qbench,qbenchplus} series have covered multiple formats on single images and image pairs, there is no existing evaluation scenario for quality comparison \textcolor{darkblue}{\textbf{beyond two images}}. 
To complement this evaluation setting, we construct the \mic. Specifically, the \mic~contains 2,000 multi-choice questions (MCQ) comparing the quality or related attributes among a group of three or four images (\textit{each half}), in which over half of the questions are {\textit{Which}} questions (\textit{e.g. which image has highest clarity?}). Aiming to extract an image with a specific quality-related appearance from a group, \textit{Which} questions are the most important questions related to image comparison. Despite \textit{Which} questions, the \mic~also includes \textit{Yes-or-No} questions and other question types (\textit{What/How/How-Many, etc}) to provide a holistic benchmark setting for multi-image quality comparison.

In summary, we conduct a systematical study towards \textcolor{deepgreen}{\textit{open-ended}} visual quality comparison. Our contributions can be summarized as three-fold:

\begin{enumerate}[itemsep=0pt, topsep=0pt, partopsep=0pt, parsep=0pt]

\item We construct the first instruction-tuning \textbf{dataset} for visual quality comparison, the Co-Instruct-562K. With data collected from two ``\textit{weak supervisors}'', \textbf{Merge2Compare} (LLM-merged comparisons) and \textbf{Teach2Compare} (GPT-4V pseudo-labeled comparisons), our public dataset significantly expands the capabilities of open-source LMMs on visual comparative settings.

\item We propose the most capable \textbf{model} for open-ended visual comparison, the \ours. With image-text interleaved input format and fine-tuned with the Co-Instruct-562K dataset, it significantly outperforms existing methods (and even GPT-4V) in multiple open-ended visual comparison tasks. With open weights, it allows for broader application than proprietary models.

\item We construct the \textbf{benchmark}, the \mic, as the first benchmark to evaluate LMMs for quality comparison on multiple images (more than two). It covers 2,000 diverse-type open-range MCQs related to visual quality comparison among three or four images. The \mic~contributes to more holistic evaluation studies on the visual quality comparison problem.
\end{enumerate}

\section{Related Works}
\label{sec:intro}

\subsection{Visual Quality Comparison}
Visual quality comparison (\textit{especially} paired comparison) is a widely used subjective quality assessment methodology, serving as the most reliable way to collect human opinions~\cite{mantiuk2012comparison}. However, when the number of images increases, the experiments become infeasible because of the exponential growth of pairwise comparisons~\cite{bt2002methodology}. While many active sampling methods have been proposed to reduce the number of pairs~\cite{ye2014active,li2018hybrid,mikhailiuk2021active}, they are computationally expensive and unsuitable for large-scale experiments. Despite subjective studies, learning to rank quality is widely proven as effective by many objective approaches~\cite{liu2017rankiqa,ma2017dipiq,golestaneh2021no,zhang2021uncertainty,wu2022fastvqa,wu2023dover,zhang2023blind}. Nonetheless, they typically only predict a scalar score or a binary judgement for overall comparison, limiting their ability to provide meaningful feedbacks into specific types of distortions or preferences.

\subsection{LMMs for Visual Quality Evaluation}

Several recent studies have explored LMMs for visual quality evaluation. The Q-Bench~\cite{wu2024qbench} proposes a holistic benchmark for LMMs on low-level perception (\textit{quality-related MCQs}), description (\textit{quality-related captioning}) and assessment (\textit{predicting scores}). Following this path, Q-Instruct~\cite{q-instruct} advances the ability of LMMs with a large-scale human-annotated dataset, and Q-Align~\cite{qalign} designs a text-guided instruction tuning for score predictions and outperforms non-LMM approaches. However, these explorations are based on single images and have not covered comparative settings. While most recent benchmarks~\cite{qbenchplus,2afc} suggest that open-source LMMs trained with single image datasets cannot perform well on comparative settings, to bridge this gap, we collect the first instruction tuning dataset to teach LMMs to compare visual quality, the Co-Instruct-562K, and our model significantly improves the ability of open-source LMMs on comparative settings, moving a step forward on the basis of existing explorations.



\begin{table*}[ht]
\vspace{-1.6em}
\centering
\renewcommand\arraystretch{1.17}
\renewcommand\tabcolsep{4pt}
\caption{Statistics of our Co-Instruct-562K dataset. `\#' denotes \textit{``the number of''}.}
\vspace{-0.66em}
\resizebox{\linewidth}{!}{\begin{tabular}{l:c:c:c:c:c}
\hline
        \textit{Subsets}          &  \textcolor{darkgray}{Q-Instruct-200K}~\cite{q-instruct}       & \textbf{Merge2Compare}        & \textbf{Teach2Compare}-\textit{general} & \textbf{Teach2Compare}-\textit{Q\&A}  & All  \\ \hline
Instruction Type & Detailed Reasoning, Question Answering & Detail Reasoning & Detail Reasoning & Question Answering & -- \\ \hdashline
 \# Total Images & \multicolumn{2}{c:}{19K \textit{(shared, both using Q-Pathway images)}} & \multicolumn{2}{c:}{9K \textit{(using shared images)}} & 28K \\ \hdashline 
 \# Total Data Items & 202K & 100K & 30K & 230K & 562K  \\ \hdashline
 \# Single Images & 202K & 0 & 0 & 0 & 202K \\
 \# Image Pairs & 0 & 70K & 18K & 134K & 222K  \\
 \# Groups of Three  &  0 & 20K & 6K & 51K & 77K \\
  \# Groups of Four  &  0 & 10K & 6K & 45K & 61K \\
\hline

\hline
\end{tabular}}
\label{tab:dataset_statistics}
\vspace{-2.2em}
\end{table*}
\section{Data Construction}
\label{sec:db}

In this section, we elaborate on the construction process of the Co-Instruct-562K dataset. Though human annotation is the most direct approach to collect data, as is widely acknowledged~\cite{li2018hybrid,ye2014active}, acquiring sufficient comparative data on a large set of images demands a markedly increased volume of human annotations~\cite{pipal,pieapp} in contrast to gathering opinions on the same set of individual images. To avoid this unbearable cost, we propose an alternative data construction strategy \textit{without additional human annotation}, by following three key principles:

\begin{enumerate}[itemsep=0pt, topsep=0pt, partopsep=0pt, parsep=0pt]
    \item \textit{Convert}: Utilize reliable information from existing datasets.
    \item \textit{Learn-from-Model}: Leverage the verified capabilities of models.
    \item \textit{Co-Instruct}: Collecting diverse subsets that complement each other.
\end{enumerate}

Under the principles, we collect two different subsets for instruction tuning: \textbf{Merge2Compare} (Sec.~\ref{sec:merge2}), which \textit{converts} information from human quality descriptions on single images, and the ability of single-modal LLMs on comparing and analyzing texts; and \textbf{Teach2Compare} (Sec.~\ref{sec:teach2}), which leverages the verified ability of GPT-4V on comparing images. Finally, we discuss how the two subsets complement each other (Sec.~\ref{sec:rationale2co}) under a \textit{co-instruct} scheme.

\subsection{{Merge2Compare}.}

\label{sec:merge2}

\begin{figure*}[h]
    \vspace{-0.3em}
    \centering
    \includegraphics[width=\linewidth]{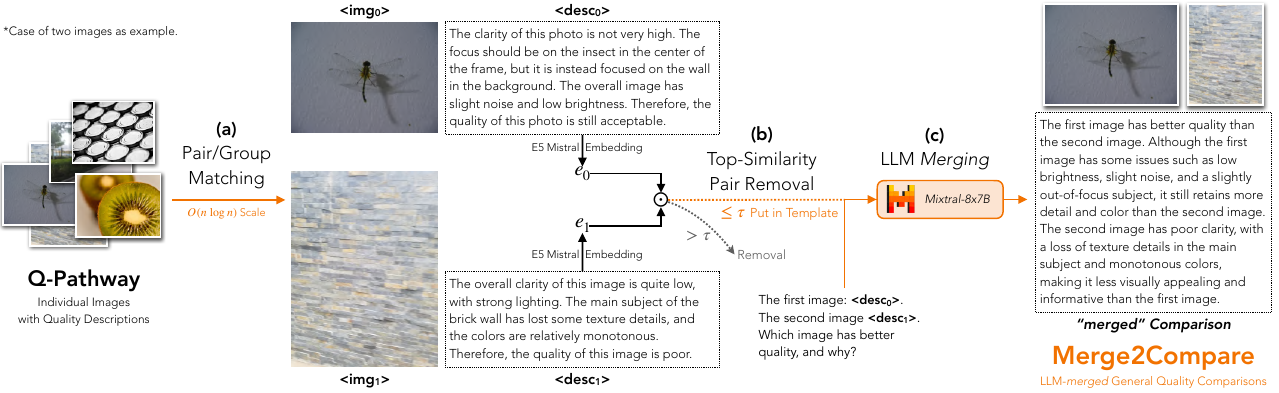}
    \vspace{-1.8em}
    \caption{The pipeline of constructing \textbf{Merge2Compare}: images are first matched into groups \textbf{(a)}, and then filtered via top-similarity removal \textbf{(b)}. After filtering, the single image quality descriptions are \textit{merged} \textbf{(c)} into comparisons by the LLM~\cite{mixtral}.}
    \label{fig:merge2compare}
    \vspace{-1.8em}
\end{figure*}

In this part, we define the construction process for \textbf{Merge2Compare} includes three steps: 1) pair/group matching; 2) top-similarity pair removal; and 3) LLM \textit{merging}. An examplar illustration of the process is shown in Fig.~\ref{fig:merge2compare}.

\paragraph{Step 1: Pair/Group Matching (Fig.~\ref{fig:merge2compare}(a)).} To best utilize information from existing single image quality descriptions, following the empirical rule to sample $O(n\log n)$ pairwise combinations to effectively rank among all individual items in a set~\cite{nlogn,pairwiseranking}, we randomly sample 81K image pairs from all 19K images in the Q-Pathway. Despite pairs, we further sample 27K groups with three images and 18K groups with four images to cover the scenarios of more images.

\paragraph{Step 2: Top-Similarity Pair Removal (Fig.~\ref{fig:merge2compare}(b)).} The effectiveness of the \textit{merge} comes from the differences among descriptions, \textit{e.g.} between \textit{The quality is acceptable} for {\tt <img$_\texttt{0}$>} and \textit{The quality is poor} for {\tt <img$_\texttt{1}$>}. However, if descriptions in a pair/group contains almost the same information (\eg both images with \textit{The clarity is good, but lighting is dark}), the \textit{merged} comparisons will lack enough information or even with false predictions. Henceforth, we use E5-Mistral~\cite{e5mistral} text embedding model to compute similarities among descriptions, and remove if \textit{any} high-similarity description pairs exist in the group. After removal, 70K image pairs (86\% of initial samples), 20K groups of three (74\% of initial) and 10K groups of four (55\% of initial) are preserved and fed into the LLM for \textit{merging}.

\paragraph{Step 3: LLM \textit{Merging} (Fig.~\ref{fig:merge2compare}(c)).} The key step for the \textbf{Merge2Compare} is to prompt LLMs to convert the single image evaluations to comparative texts. Specifically, following many existing practices~\cite{llava,minigpt4,lamm}, we put the descriptions as alternates of images in the context. Denote the description for image {\tt <img$_\texttt{i}$>} as {\tt <desc$_\texttt{i}$>}, the user query for LLM \textit{merging} is formulated as follows:\\
(Pairs) {{\tt The first image: {<desc$_\texttt{0}$>} The second image: {<desc$_\texttt{1}$>} \\}}
{{\tt~~Which image has better quality, and why?\\}}
(Groups of Three/Four) {{\tt $\{$The $K_{i+1}$ image: {<desc$_\texttt{i}$>} $|_{i=0}^{N-1}\}$ \\}}
{{\tt~~Please rank the quality of the images and justify your rankings.}}\\
where $K_i$ represents the ordinal form of $i+1$, \eg $K_1$ is {\tt first}, $K_2$ is {\tt second}.

The \textit{merged} comparisons are \textbf{overall comparisons with reasonings} (Fig.~\ref{fig:merge2compare} \textit{right}). To validate their reliability, we conducted a human examination on 250 random samples from \textbf{Merge2Compare} mixed with merged comparisons from 250 random removed groups. The correctness rate in \textbf{Merge2Compare} is 96\%, while it is only 72\% in the removed groups, demonstrating the effects of the top-similarity removal process in ensuring the quality of training data.

\begin{figure*}[h]

    \centering
    \includegraphics[width=0.96\linewidth]{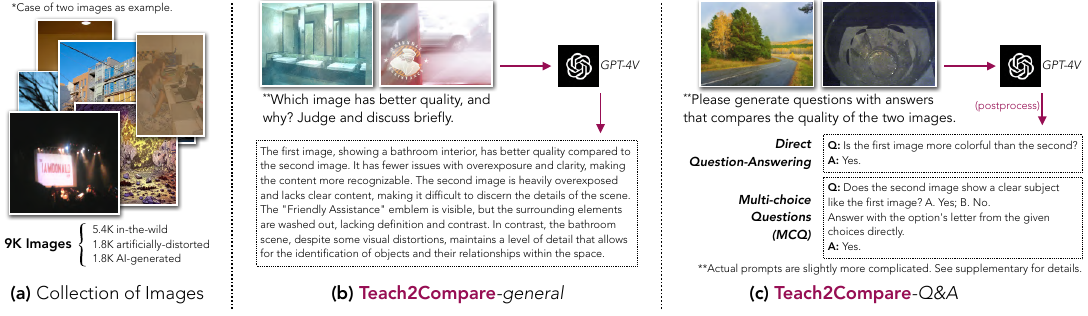}
    \vspace{-1em}
    \caption{The pipeline of constructing \textbf{Teach2Compare}: 9K diverse images are collected and matched into 30K groups \textbf{(a)}. The groups are then fed to GPT-4V to obtain \textit{general} quality comparisons \textbf{(b)} and \textit{question-answering} \textbf{(c)} related to quality comparisons.}
    \label{fig:teach2compare}
    \vspace{-1.8em}
\end{figure*}

\subsection{{Teach2Compare}.}
\label{sec:teach2}

Given existing evaluations~\cite{2afc,qbenchplus} suggesting that GPT-4V is decent at comparing visual quality (Tab.~\ref{tab:qpaira1}/\ref{tab:qpaira2}/\ref{tab:2afclmm}), we propose to collect GPT-4V responses as pseudo labels for quality comparison. As shown in Fig.~\ref{fig:teach2compare}, we collect diverse unlabeled images and feed them to GPT-4V with different prompts to obtain \textbf{Teach2Compare}-\textit{general} (overall quality comparison) and \textbf{Teach2Compare}-\textit{Q\&A} (question-answer pairs related to quality comparison). Details as follows.

\paragraph{Collection of Images (Fig.~\ref{fig:teach2compare}(a)).} For \textbf{Teach2Compare}, we collect 9K images from various sources to cover different quality concerns and visual appearances: \textbf{1)} 5.4K \textit{in-the-wild} images from YFCC-100M~\cite{yfcc} database; \textbf{2)} 1.8K images \textit{artificially-distorted} images from COCO~\cite{coco_caption} (with 15 types of distortions via ImageCorruptions~\cite{imagecorruptions}) and KADIS-700K~\cite{kadid} (25 types of distortions); \textbf{3)} 1.8K \textit{AI-generated} images from ImageRewardDB~\cite{imagereward}. These 9K diverse unlabeled images are further grouped into 18K pairs, 6K groups of three, and 6K groups of four, for GPT-4V to provide pseudo labels under two formats, as follows.

\paragraph{\emph{\textbf{Teach2Compare}}-general (Fig.~\ref{fig:teach2compare}(b)).} Similar to \textbf{Merge2Compare}, the \textit{general} subset also consists of overall comparison with reasonings. Specifically, we substitute the {\tt <desc$_\texttt{i}$>} in the \textbf{Merge2Compare} prompt template to respective real images {\tt <img$_\texttt{i}$>} to feed to GPT-4V. After collection, we also conduct a similar 250-sample spot check on the output pseudo labels, which reports around 94\% correctness. Though slightly less accurate than \textbf{Merge2Compare} (96\%), examiners also report that GPT-4V labels contain more \textbf{content information} which has been observed to enhance quality understandings of models~\cite{fastervqa,vsfa,sfa}. The two subsets are expected to complement each other for better learning outcomes.

\paragraph{\emph{\textbf{Teach2Compare}}-Q\&A (Fig.~\ref{fig:teach2compare}(c)).} Despite general comparisons, for GPT-4V, we also collect a specific subset to improve LMM ability on responding to open-range questions. To achieve this, we first list reference aspects (\textit{clarity, lighting, color, etc}) and then ask GPT-4V to generate questions (and respective correct answers, and false answers for the questions) on comparing these aspects among the images. After removing failed generations, we obtain 230K question-answers from 30K image groups. These question-answers are converted to both direct question-answering and multi-choice questions (as A-OKVQA\cite{AOKVQA}) for training. 


\begin{figure*}[h]
\vspace{-0.5em}
    \centering
    \includegraphics[width=\linewidth]{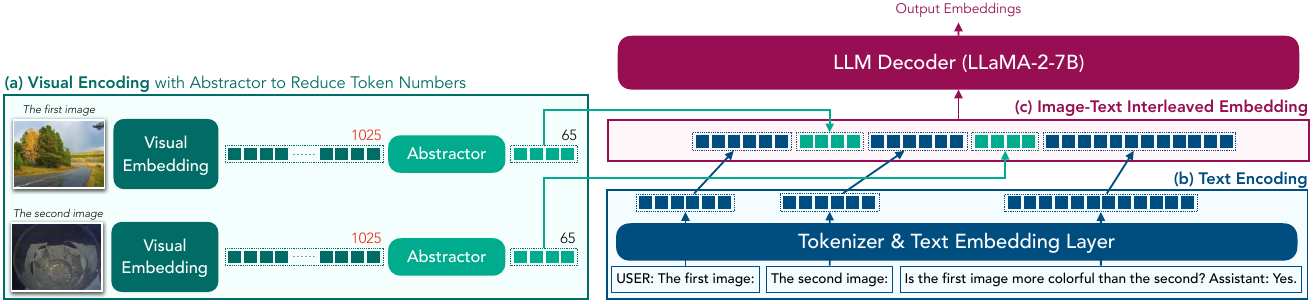}
    \vspace{-1.8em}
    \caption{\textbf{The structure of \ours.} \textbf{(a)} Images are encoded by visual embedding layers and then passsed through an abstractor module to reduce token numbers, and then \textbf{(c)} fused with text embeddings into under the image-text interleaved format.}
    \label{fig:coinstructmodel}
    \vspace{-1.4em}
\end{figure*}

\subsection{Rationale of Combinations.}
\label{sec:rationale2co}

As discussed in principle 3, our motivation is to collect subsets that can complement each other. This complementarity is reflected in the following two aspects. Firstly, in terms of general comparisons, \textbf{Merge2Compare} has higher accuracy but lacks fine-grained comparison (excluded by \textit{Top-similarity Pair Removal}), while \textbf{Teach2Compare}-\textit{general}, although slightly less accurate, offers more diverse scenarios and includes content information as background. Joint training of both contributes to a more comprehensive quality comparison by our model. Additionally, \textbf{Teach2Compare} includes a unique \textit{Q\&A} subset, which significantly enhances the model's ability to answer open-range questions.


\section{The \ours~Model}
\label{sec:mtd}

In this section, we discuss the proposed model, \ours. Specifically, we have made two non-trivial adaptations for the multi-image comparative setting:

\paragraph{Visual Token Reduction (Fig.~\ref{fig:coinstructmodel} (a)).} Most state-of-the-art LMMs~\cite{improvedllava,bakllava,internlmxcomposer2} have adopted the simple projector that keeps a large number of tokens (\textit{e.g.} 1,025). This structure is not friendly to multi-image scenarios: passing only two images will exceed the max length of LLaVA~\cite{llava} (2,048), and four images will exceed the context window of LLaMA-2~\cite{llama2} (4,096). Thus, we adopt another widely-used abstractor~\cite{iblip,mplugowl,mplug2} structure to reduce token numbers before feeding the visual embeddings to LLM, so as to easily adapt to multi-image scenarios.

\paragraph{Image-text Interleaved Format (Fig.~\ref{fig:coinstructmodel} (c)).} Typical single-image instruction tuning usually does not care about ``\textit{position of images}''. Most approaches~\cite{minigpt4,improvedllava,iblip} directly pile all images before texts ({\tt \textcolor{darkblue}{<img$_\texttt{0}$>}\textcolor{darkblue}{(<img$_\texttt{1}$>$\dots$)}<text>}). Under this piling, multiple images are not separates and LMMs might confuse the information from different images and fail to compare well (see baseline result in Fig.~\ref{fig:qualitativepaira2}). To solve this, we propose an image-text interleaved format for multi-image training, that each image is started with explicit text to identify its nominal:

{{\noindent \tt User: The first image: \textcolor{darkblue}{<img$_\texttt{0}$>}~The second image: \textcolor{darkblue}{<img$_\texttt{1}$>}~($\dots$)~<query>\\}}{{\tt Assistant: <response>}}

In our experiments, we demonstrated that this interleaved format significantly enhances the performance of \ours~(Tab.~\ref{tab:ablinterleave}), notably better than using learnable special tokens ({\tt <img\_st>} and {\tt <img\_end>}) to divide images.

\begin{figure*}[h]
    \centering
    \includegraphics[width=\linewidth]{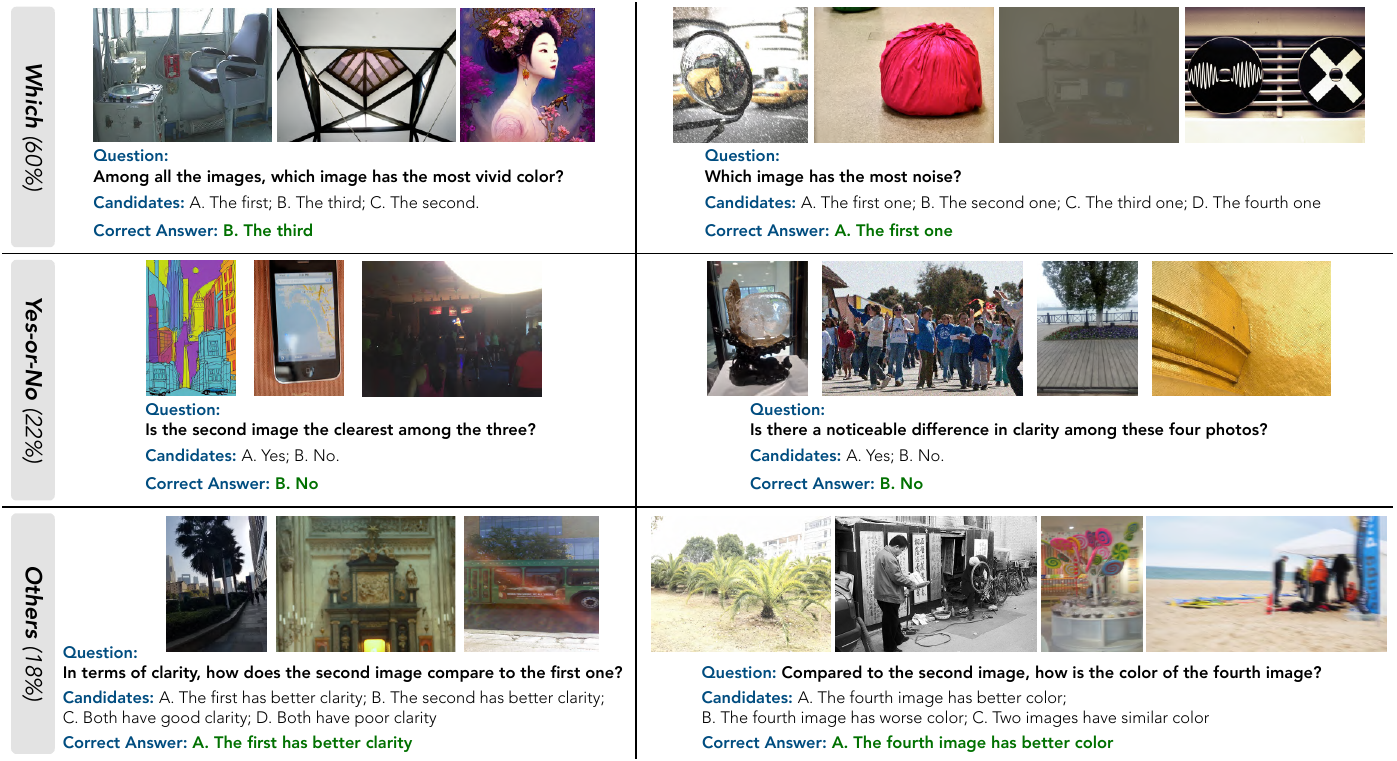}
    \vspace{-1.8em}
    \caption{\textbf{Dataset Card of \mic}, made up of (a) \textit{Which} questions (60\%), (b) \textit{Yes-or-No} questions (22\%), and (c) \textit{Other} types of questions (18\%) on three/four images.}
    \label{fig:micmcq}
    \vspace{-1.6em}
\end{figure*}

\section{The \mic}
\label{sec:micbench}

In this section, we discuss the proposed \mic~to cover the open-ended evaluation settings on \textcolor{darkblue}{groups of three or four images}, as a complementary of existing evaluation settings (Sec.~\ref{sec:existingbench}). It contains 2,000 groups of \textit{open-range} questions equipped with multiple candidates, with details elaborated as follows:

\paragraph{Sourcing Diverse Image Groups.} To improve the diversity of the benchmark, in the {\mic}, we sample image groups from two sources: \textbf{(1)} 400 groups of three and 400 groups of four from the images in LLVisionQA~\cite{wu2024qbench}, which are originally sourced from 9 datasets~\cite{agiqa3k,zhang2023subjective,spaq,koniq,paq2piq,clive,kadid,livemultipledistortions,coco_caption}; \textbf{(2)} 600 groups of three and 600 groups of four on 1,000 random images sampled from unlabeled databases~\cite{yfcc,kadid,coco_caption,imagereward} (\textit{zero overlap with training-set images}). With in-total 2,000 groups, the \mic~contains a wide variety of quality issues and low-level appearances, providing a non-biased evaluation on quality comparison.

\paragraph{Evaluation Form: Multi-choice Questions (MCQs).} As the most popular evaluation form for LLM/LMM benchmarks~\cite{mmlu,mmmu,mmbench,wu2024qbench}, multi-choice question (MCQ)  is adopted as the evaluation form of \mic. As is shown in Fig.~\ref{fig:micmcq}, each image group is associated with a expert-crafted question that compare quality or related attributes among the images. Despite common question types (\textit{Yes-or-No/What/How, etc}), the \mic~also introduces a special type of question, the \textbf{\textit{Which}} questions (Fig.~\ref{fig:micmcq}(a)), to cover this common type of human query on comparison. In total 10 human experts participate in annotating the \mic, and the answer of each MCQ is cross-examined by another expert. Similar as existing benchmarks~\cite{mmbench,wu2024qbench}, \mic~is further divided into a \textit{dev} set (1,004) for method development (\textit{answers will be public}), and a \textit{test} set (996) to evaluate performance of LMMs (\textit{answers will be hidden from public}).


\section{Evaluation}

\subsection{Implementation Details}

The \ours~is fine-tuned after the released checkpoint of mPLUG-Owl2~\cite{mplug2}, with LLaMA-2~\cite{llama2} as LLM and CLIP-ViT-L14~\cite{clip} as visual embedding module. Images are padded to square and then resized to $448\times448$ before fed into the model. The learning rate is set as $2e$-$5$, with two epochs under batch size $192$. The final checkpoint is used for evaluation. To avoid over-fitting, only \textit{dev} subsets of evaluation datasets are used to choose best training hyper-parameters, where the final reported results are from non-overlapped \textit{test} subsets. All parameters are updated during training, costing in total 25 hours on 8*NVIDIA A100 GPUs.

\subsection{Baseline Models}

We choose five open-source recent state-of-the-art LMMs that supports multi-image inputs to compare with: LLaVA-v1.5-13B~\cite{improvedllava}, InternLM-XComposer2~\cite{internlmxcomposer2}, BakLLaVA~\cite{bakllava}, EMU2-Chat~\cite{emu2}, mPLUG-Owl2~\cite{mplug2} (\textit{baseline of} \ours). Additionally, we also compare with three well-recognized proprietary close-source models:  Qwen-VL-Max, Gemini-Pro, and GPT-4V (\textit{teacher of} \ours).

\subsection{Results on Existing Evaluation Settings}
\label{sec:existingbench}

Despite the \mic~(Sec.~\ref{sec:micbench}), we also evaluate the proposed \ours~against baseline models on several existing visual quality evaluation/comparison benchmarks for LMMs. The evaluation settings and results are as follows.

\begin{table}[]
    \centering
    \vspace{-1.5em}
    \renewcommand\arraystretch{1.22}
    \renewcommand\tabcolsep{6pt}
    \caption{Results on \textit{Q-Bench$^\text{\tt PAIR}$-A1}. \ours~is remarkably \textbf{51\%} better than the variant \textit{without comparative data}, and the only LMM that surpasses human capability. }
    \vspace{-8pt}
    \resizebox{\linewidth}{!}{\begin{tabular}{l|ccc|cc|cc|c}
    \hline
    \textbf{Sub-categories} & \multicolumn{3}{c|}{\textbf{Question Types}} & \multicolumn{2}{c|}{\textbf{Low-level Concerns}} & \multicolumn{2}{c|}{\textbf{Pairwise Settings}} & \multirow{3}{*}{{\textit{Overall$\uparrow$}}} \\ \cdashline{1-8}
        \multirow{2}{*}{\textbf{Model}}  & \multirow{2}{*}{\textit{Yes-or-No$\uparrow$}}& \multirow{2}{*}{\textit{What$\uparrow$}} & \multirow{2}{*}{\textit{How$\uparrow$}} & \multirow{2}{*}{\textit{Distortion$\uparrow$}} & \multirow{2}{*}{\textit{Other$\uparrow$}} & \multirow{2}{*}{\textit{Compare$\uparrow$}}  &\multirow{2}{*}{\textit{Joint$\uparrow$}}  \\
        &&&&&&& \\ \hline
    \textit{random guess accuracy} & 50.00\% & 32.03\% & 33.16\% & 38.95\% & 41.95\% & 38.69\% & 43.70\% & 39.82\% \\ \cdashline{1-9}
          (\texttt{Sep/2023}) LLaVA-v1.5-13B & 57.34\% & {47.45\%} & 49.13\% & 49.01\% & {59.51\%} & {52.06\%} & 52.00\% & 52.05\%  \\
         (\texttt{Oct/2023}) BakLLava & 60.09\% & {45.42\%} & {50.86\%} & {53.09\%} & {58.82\%} & {54.52\%} & {55.55\%} & {52.75\%}\\
         (\texttt{Nov/2023}) mPLUG-Owl2 \textit{(baseline of \ours)} & 58.07\% & 36.61\% & 48.44\% & 47.74\% & 51.90\% & 45.73\% & {60.00\%} & 48.94\% \\
         (\texttt{Dec/2023}) Emu2-Chat & 51.94\% & 29.78\% & {53.84\%} & 42.01\% & 55.71\% & 46.26\% & 49.09\% & 47.08\% \\
        
        (\texttt{Feb/2024}) InternLM-XComposer2-VL  & 71.81\% & 58.64\% & 62.28\% & 65.77\% & 63.67\% & 64.34\% & \textbf{68.00\%} & 65.16\% \\
        \hdashline
        {Qwen-VL-Max} \textit{(Proprietary)}  & 67.65\% & 67.56\% & 65.35\% & 69.09\% & 61.18\% & 68.65\% & 61.29\% & 66.99\%  \\
        {Gemini-Pro} \textit{(Proprietary)} & 65.78\% & 56.61\% & 56.74\% & 60.42\% & 60.55\% & 60.46\% & 60.44\% & 60.46\%  \\ 
         {GPT-4V} \textit{(Proprietary, teacher of \ours)} & \textbf{79.75\%} & \textbf{69.49\%} & \gb{84.42\%} & \textbf{77.32\%} & \textbf{79.93\%} & \textbf{81.00\%} & \textbf{68.00\%} & \textbf{78.07\%}  \\ \hline
         \textit{Non-expert \textit{Human}} & 78.11\% & 77.04\% & 82.33\% & 78.17\% &  77.22\% & 80.26\% & 76.39\% & {80.12\%}  \\ \hline
         \textit{without Multi-image Comparative Data}   & {60.24\%} & {47.46\%} & 48.78\% & {52.81\%} & 53.97\% & 51.42\% & {59.11\%} & {53.15\%}  \\ \hdashline
         \ours~(Ours) & \gb{86.50\%} & \gb{72.20\%} & \textbf{79.23\%} & \gb{80.00\%} & \gb{80.62\%} & \gb{81.91\%} & \gb{74.22\%} & \gb{80.18\%} \\ \hline

   \end{tabular}}
    \label{tab:qpaira1}
    \vspace{-1.5em}
\end{table}



\noindent \textbf{Q-Bench$^\text{\tt PAIR}$-A1}\cite{qbenchplus} is a benchmark for visual quality comparison with 1,999 expert-crafted \textcolor{deepgreen}{\textit{open-range}} quality-related MCQs on \textit{image pairs}. In Tab.~\ref{tab:qpaira1}, we compare \ours~against existing open-source and proprietary models on this benchmark. \ours~shows far superior accuracy than open-source LMMs: it is \textbf{64\%} better than its baseline (mPLUG-Owl2), \textbf{51\%} better than the variant without our multi-image subsets (\textbf{Merge2Compare} and \textbf{Teach2Compare}), and also 23\% better than the best of them. It also outperforms Qwen-VL-Max and Gemini-Pro by a large margin (21\%/33\%). Additionally, though its all MCQ training data are from GPT-4V, the student (\ours) still outperforms its teacher on this MCQ evaluation set by notable \textbf{2.7\%}, suggesting the effectiveness of the collaborative teaching strategy. Our model is also \textit{the only LMM} that surpasses the accuracy of a non-expert human (\textit{esp.} on Compare subset) in this benchmark, strongly supporting the meaningful vision of using models to relieve human labors on real-world visual quality comparisons in the future.

\begin{table*}[htbp]
\vspace{-1.5em}
        \centering
    \renewcommand\arraystretch{1.25}
    \renewcommand\tabcolsep{4.5pt}
        \caption{Results on \textit{Q-Bench$^\text{\tt PAIR}$-A2}. $P_i$ denotes frequency for score $i$ (score in $[0,2]$). While slightly inferior to GPT-4V, \ours~has significantly improved over both the variant \textit{without comparative data} (\textbf{+31\%}), especially for \textbf{Precision} metric (\textbf{+59\%}).}
        \vspace{-8pt}
    \resizebox{\linewidth}{!}{\begin{tabular}{l|cccc|cccc|cccc|c}
    \hline
        \textbf{Dimensions} & \multicolumn{4}{c|}{\textbf{Completeness}} & \multicolumn{4}{c|}{\textbf{Precision}} & \multicolumn{4}{c|}{\textbf{Relevance}} & \multirow{2}{*}{\textit{Sum.$\uparrow$}} \\ \cdashline{1-13}
        \textbf{Model} & $P_0$ & $P_1$ & $P_2$ & \textit{score$\uparrow$}   &  $P_0$ & $P_1$ & $P_2$ & \textit{score$\uparrow$}   & $P_0$ & $P_1$ & $P_2$  & \textit{score$\uparrow$} \\ \hline
         (\texttt{Sep/2023}) LLaVA-v1.5-13B  & 18.77\% & 73.44\% & 7.79\% & {0.89} & 34.66\% & 38.72\% & 26.62\% & {0.92} & 1.02\% & 34.59\% & 64.39\% & {1.63} & {3.44} \\
        (\texttt{Oct/2023}) BakLLava & 29.46\% & 59.77\% & 10.57\% & 0.80 & 40.0\% & 38.08\% & 21.33\% & 0.80 & 2.26\% & 15.06\% & 82.04\% & \underline{1.79} & 3.40\\

         (\texttt{Nov/2023}) mPLUG-Owl2 \textit{(baseline)} & 19.43\% & 65.54\% & 14.45\% & {0.94} & 30.94\% & 43.71\% & 24.63\% & {0.92} & 3.79\% & 26.94\% & 68.28\% & {1.63} & {3.50}\\

        (\texttt{Dec/2023}) Emu2-Chat& 41.25\% & 54.33\% & 4.42\% & 0.63 & 38.11\% & 36.41\% & 25.48\% & {0.87} & 4.12\% & 38.61\% & 57.27\% & 1.53 & 3.03 \\ 
        (\texttt{Feb/2024}) InternLM-XComposer2-VL  & 13.20\% & 72.17\% & 14.13\% & 1.00 & 31.28\% & 42.13\% & 25.77\% & 0.93 & 1.60\% & 24.17\% & 72.93\% & 1.70 & 3.64 \\
        \hdashline

Qwen-VL-Max \textit{(Proprietary)} & 11.64\% & 54.08\% & 34.08\% & 1.22 & 24.26\% & 39.15\% & 36.22\% & 1.11 & 2.533\% & 10.97\% & 85.64\% & 1.82 & 4.16 \\ 
Gemini-Pro \textit{(Proprietary)} & 18.22\% & 44.48\% & 36.84\% & 1.18 & 34.13\% & 37.95\% & 27.02\% & 0.92 & 0.67\% & 5.91\% & 92.22\% & 1.90 & 4.00  \\ 
{GPT-4V} \textit{(Proprietary, teacher of \textbf{Ours})} &  4.09\% & 31.82\% & 64.09\% & \gb{1.60} & 10.44\% & 45.12\% & 44.44\% & \gb{1.34} & 0.18\% & 1.69\% & 96.35\% & \textbf{1.94} & \gb{4.89} \\ \hline
\textit{w/o Multi-Image Comparative Data} & 15.25\% & 65.76\% & 18.32\% & {1.02} & 39.44\% & 40.18\% & 19.62\% & 0.79 & 0.09\% & 9.86\% & 89.02\% & {1.87} & {3.69}\\ \hdashline
\ours~(Ours) & 4.04\% & 31.55\% & 63.55\% & \textbf{1.58} & 13.68\% & 43.68\% & 41.37\% & \textbf{1.26} & 0.0\% & 0.44\% & 98.22\% & \gb{1.96} & \textbf{4.82} \\
        \hline
    \end{tabular}}
    \vspace{-1.6em} 
    \label{tab:qpaira2}
\end{table*}

\begin{figure*}[h]
\vspace{-2.2em}
    \centering
    \includegraphics[width=\linewidth]{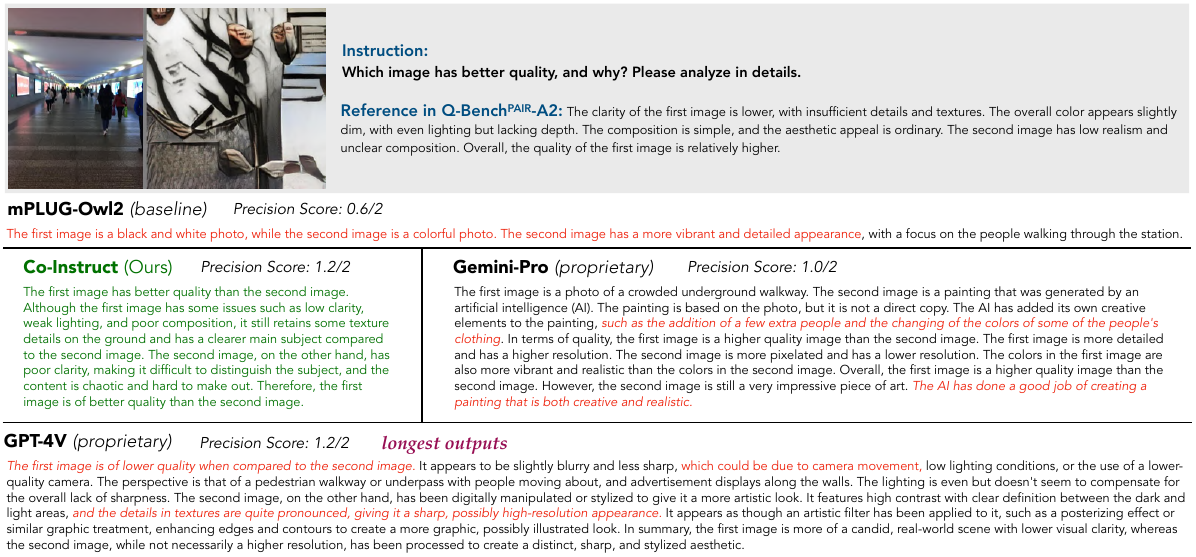}
    \vspace{-1.8em}
    \caption{\textbf{Qualitative Visualization on}  \textit{Q-Bench$^\text{\tt PAIR}$-A2}. GPT-4V gives longest outputs and achieves high precision score even if it includes incorrect information.}
    \label{fig:qualitativepaira2}
    \vspace{-1.8em}
\end{figure*}

\noindent \textbf{Q-Bench$^\text{\tt PAIR}$-A2} is a benchmark setting for general and detailed visual quality comparison \textcolor{deepgreen}{\textit{with detailed reasonings}} on \textit{image pairs}. Consisting of 499 image pairs, it employs GPT to evaluate LMM responses against the \textit{golden} expert-labeled comparisons on \textbf{Completeness}, \textbf{Precision}, 
and \textbf{Relevance}. As listed in Tab.~\ref{tab:qpaira2}, the \ours~can remarkably improve the \textbf{Completeness} (+57\%) and \textbf{Precision} (+59\%) of the comparative outputs than the \textit{w/o comparative data} version, but still falls a little bit behind GPT-4V on the quantitative metrics. This might be because outputs from GPT-4V are more than \textit{twice as long} as \ours~outputs, while GPT evaluation used here is observed~\cite{alpaca_eval} to be in favor of {longer text outputs}. To further analyze this potential bias, we qualitatively visualize the result of different LMMs in Fig.~\ref{fig:qualitativepaira2}. As shown in the figure, the baseline open-source LMM even confuses the information from the two images, and Gemini-Pro makes rather poor detailed reasonings. For GPT-4V, it generates the \textcolor{purple}{\textbf{\textit{longest outputs}}} among all LMMs, which might be the reason that it gets a relatively high precision score even its outputs are not totally correct. In short, the capability of \ours~in reasoning-related comparisons can match that of GPT-4V, while significantly surpassing other existing LMMs.


\begin{table*}\small
    \centering
    \vspace{-1.8em}
    \renewcommand\arraystretch{1.17}
    \renewcommand\tabcolsep{4.5pt}
        \caption{Results on \textit{2AFC-LMM}. $\kappa$ denotes binary judgment consistency while swapping \textit{first image} and \textit{second image}; $\rho$ denotes Pearson's linear correlation.}
        \vspace{-8pt}
    \resizebox{\linewidth}{!}{\begin{tabular}{l|cc|cc|cc|cc|cc|cc|cc}
    \hline
    \textbf{Dataset} & \multicolumn{2}{c|}{\textbf{CSIQ}} & \multicolumn{2}{c|}{\textbf{MM21}} & \multicolumn{2}{c|}{\textbf{KADID-10k}} & \multicolumn{2}{c|}{\textbf{LIVEC}} & \multicolumn{2}{c|}{\textbf{KonIQ-10k}} & \multicolumn{2}{c|}{\textbf{SPAQ}} & \multicolumn{2}{c}{\textit{weighted avg.}} \\
    \textbf{Model} & $\kappa$ & $\rho$ & $\kappa$ & $\rho$ & $\kappa$ & $\rho$ & $\kappa$ & $\rho$ & $\kappa$ & $\rho$ & $\kappa$ & $\rho$ & $\kappa$ & $\rho$ \\ \hline
     (\texttt{Aug/2023}) IDEFICS-Instruct-9B  & 0.206 & 0.570 & 0.337 & 0.338 & 0.202 & 0.552 & 0.323 & 0.492 & 0.251 & 0.479 & 0.330 & 0.474  & 0.286 & 0.470 \\
             (\texttt{Sep/2023}) LLaVA-v1.5-13B  & 0.483 & 0.423 & 0.356 & 0.149 & 0.310 & 0.137 & 0.273 & 0.162 & 0.262 & 0.403 & 0.291 & 0.156 & 0.302 & 0.224  \\
        (\texttt{Oct/2023}) BakLLava & 0.356  & 0.235 & 0.337 & 0.244 & 0.245 & 0.166 & 0.296 & 0.159 & 0.185 & 0.217 & 0.274 & 0.146 & 0.261 & 0.185 \\

         (\texttt{Nov/2023}) mPLUG-Owl2 \textit{(baseline)} & 0.435 & 0.627 & 0.378 & 0.306 & 0.402 & 0.443 & 0.375 & 0.441 & 0.386 & 0.417 & 0.362 & 0.356 & 0.460 &  0.397 \\

        (\texttt{Feb/2024}) InternLM-XComposer2-VL & 0.800 & 0.527 & 0.688 & 0.377 & 0.600 & 0.552 & 0.600 & 0.516 & 0.825 & 0.581 & 0.700 & 0.755 & 0.705 & 0.567  \\
        \hdashline
Qwen-VL-Max \textit{(Proprietary)} & 0.540 & 0.418 & 0.497 & 0.304 & 0.625 & 0.406 & 0.578 & 0.544 & 0.631 & 0.610 & 0.592 & 0.718 & 0.592 & 0.540 \\ 
Gemini-Pro \textit{(Proprietary)} & 0.672 & 0.527 & 0.604 & 0.377 & 0.790 & 0.552 & 0.650 & 0.516 & 0.652 & 0.581 & 0.671 & 0.755 & 0.678 & 0.622 \\ 
{GPT-4V} \textit{(Proprietary, teacher of \textbf{Ours})} & 0.778 & 0.764 & 0.792 & \textbf{0.474} & 0.763 & 0.560 & 0.837 & 0.685 & 0.835 & 0.800 & 0.871 & 0.876 & 0.823 & 0.721 \\ \hline
\textit{w/o Multi-Image Comparative Data} & 0.117 & 0.650 & 0.480 & 0.392 & 0.397 & 0.466 &  0.327 & 0.432 & 0.489 & 0.512 & 0.485 & 0.397 & 0.432 & 0.449  \\ \hdashline
\ours~(Ours)& \textbf{0.800} & \textbf{0.779} & \textbf{0.852} & 0.325 & \textbf{0.829} & \textbf{0.685} & \textbf{0.872} & \textbf{0.797} & \textbf{0.883} & \textbf{0.927} & \textbf{0.881} & \textbf{0.931} & \textbf{0.864} & \textbf{0.754}   \\  \hline
    \end{tabular}}
    \vspace{-1.5em} 
    \label{tab:2afclmm}
\end{table*}

\noindent \textbf{2AFC-LMM}\cite{2afc} is a benchmark setting for general quality comparison on \textit{image pairs}. It prompts LMMs to make a \textit{two-alternative forced choice} (2AFC) on a pair of images. The maximum a posterior estimation is utilized to aggregate comparative preferences to single-image quality scores~\cite{howtocompare}.
Then, it computes Peason's linear correlation ($\rho$) between regressed scores and ground truth MOS. As shown in Tab.~\ref{tab:2afclmm},  \ours~outperforms all existing models in \textit{2AFC-LMM}, including GPT-4V. \ours~also shows very high consistency $\kappa$ while swapping two images. Among all datasets, the proposed model is only inferior on the MM21~\cite{mm21db}. Nonetheless, we observe that the \ours~has higher direct comparison accuracy than GPT-4V (\ours: \underline{55.2\%}, GPT-4V: \underline{54.4\%}, see supplementary for full results) on it, yet the dataset contains a large proportion of \textit{extremely similar pairs}, for which \ours~responds a forced choice but GPT-4V will answer \textit{``two images have similar quality''} (tie), which impacts the aggregated single-image quality scores. We hope this observation can inspire further research to design better evaluation settings for fine-grained comparisons.

\begin{table*}\small
\vspace{-1.5em}
    \centering
    \renewcommand\arraystretch{1.22}
    \renewcommand\tabcolsep{6pt}
    \caption{Results on \textit{Q-Bench$^\text{\tt SINGLE}$-A1}, proving that the comparative data (Sec.~\ref{sec:db}) can also effectively boost the capability of LMMs on single image quality evaluation. }
    \vspace{-8pt}
    \resizebox{\linewidth}{!}{\begin{tabular}{l|ccc|cccc|c}
    \hline
    \textbf{Sub-categories} & \multicolumn{3}{c|}{\textbf{Question Types}} & \multicolumn{4}{c|}{\textbf{Quadrants of Low-level Concerns}} & \multirow{3}{*}{{\textit{Overall$\uparrow$}}} \\ \cdashline{1-8}
        \multirow{2}{*}{\textbf{Model}}  & \multirow{2}{*}{\textit{Yes-or-No$\uparrow$}}& \multirow{2}{*}{\textit{What$\uparrow$}} & \multirow{2}{*}{\textit{How$\uparrow$}} & \multirow{2}{*}{\textit{Distortion$\uparrow$}} & \multirow{2}{*}{\textit{Other$\uparrow$}} & \textit{In-context}  &\textit{In-context}  \\
        &&&&&&\textit{Distortion$\uparrow$}& \textit{Other$\uparrow$} \\ \hline
    \textit{random guess accuracy} & 50.00\% & 28.48\% & 33.30\% & 37.24\% & 38.50\% & 39.13\% & 37.10\% & 37.94\% \\ \cdashline{1-9}
          (\texttt{Sep/2023}) LLaVA-v1.5-13B & 64.96\% & {64.86}\% & 54.12\% & 53.55\% & {66.59}\% & {58.90}\% & 71.48\% & 61.40\% \\
         (\texttt{Oct/2023}) BakLLava & 66.46\% & 61.48\% & 54.83\% & 51.33\% & 63.76\% & 56.52\% & {78.16}\% & 61.02\% \\
         (\texttt{Nov/2023}) mPLUG-Owl2 \textit{(baseline of \ours)} & {72.26}\% & 55.53\% & {58.64}\% & 52.59\% & {71.36}\% & 58.90\% & 73.00\% & 62.68\% \\
         (\texttt{Dec/2023}) Emu2-Chat & 70.09\% & {65.12}\% & 54.11\% & {66.22}\% & 62.96\% & {63.47}\% & 73.21\% & {64.32}\% \\
        
        (\texttt{Feb/2024}) InternLM-XComposer2-VL  & 72.44\% & 78.13\% & 67.28\% & 68.00\% & \textbf{75.65\%} & 68.15\% & {81.36\%} & 72.52\% \\
        \hdashline
        {Qwen-VL-Max} \textit{(Proprietary)}  & 73.20\% & \gb{81.02\%} & 68.39\% & 70.84\% & {74.57\%} & 73.11\% & {80.44\%} & 73.90\% \\
        {Gemini-Pro} \textit{(Proprietary)} & 71.26\% & 71.39\% & 65.59\% & 67.30\% & 73.04\% & 65.88\% & 73.60\% & 69.46\% \\ 
         {GPT-4V} \textit{(Proprietary, teacher of \ours)} & {77.72\%} & 78.39\% & 66.45\% & \textbf{71.01\%} & 71.07\% & \gb{79.36\%} & 78.91\% & \textbf{74.10\%}  \\ \hline
         \textit{Non-expert \textit{Human}} &82.48\% & 79.39\% & 60.29\% & 75.62\% & 72.08\% & 76.37\% & 73.00\% & 74.31\%  \\ \hline

         \textit{without Multi-image Comparative Data}  &  \textbf{79.38\%} & 72.23\% & 67.70\% & 68.71\% & 72.32\% & 73.97\% & \textbf{83.65\%} & 73.38\% \\ \hdashline
         \ours~(Ours) &  \gb{81.93\%} & \textbf{78.74\%} & \gb{70.16\%} & \gb{74.28\%} &  \gb{76.37\%} & \textbf{76.71\%} & \gb{84.41\%} & \gb{77.12\%} \\ \hline
   \end{tabular}}
    \vspace{-1.8em}
    \label{tab:qsinglea1}
\end{table*}

\noindent \textbf{Q-Bench$^\text{\tt SINGLE}$-A1.} Despite the comparative benchmarks above, we also evaluate \ours~on \textit{single image} MCQs from \textit{Q-Bench$^\text{\tt SINGLE}$-A1} to verify the influences of comparative tuning on single-image  quality perception. As shown in Tab.~\ref{tab:qsinglea1}, \ours~shows \textbf{5\%} improvement over the variant \textit{trained with single images only}, leading GPT-4V by 4\%, and marks the only LLM that surpasses non-expert human. These results have demonstrated the contribution of comparative training on general quality-related understanding of LMMs, and suggested that single-image quality evaluation \textit{does not conflict with} multi-image quality comparison for LMMs and can be improved together under a unified model.

\subsection{Results on \mic}

\begin{table}
    \centering
    \vspace{-1.5em}
    \renewcommand\arraystretch{1.22}
    \renewcommand\tabcolsep{7.5pt}
    \caption{Results on {\mic}. \ours~is \textbf{60\%} better than the variant \textit{without comparative data}, and also notably better than GPT-4V (+5.7\%) and human (+6.4\%).}
    \vspace{-8pt}
    \resizebox{\linewidth}{!}{\begin{tabular}{l|ccc|cc|c}
    \hline
    \textbf{Sub-categories} & \multicolumn{3}{c|}{\textbf{Question Types}} & \multicolumn{2}{c|}{\textbf{Number of Images}} & \multirow{2}{*}{{\textit{Overall$\uparrow$}}}  \\  \cdashline{1-6}
        \multirow{1}{*}{\textbf{Model}}  & \multirow{1}{*}{\textit{Yes-or-No$\uparrow$}}& \multirow{1}{*}{\textit{Which$\uparrow$}} & \multirow{1}{*}{\textit{Others$\uparrow$}}  & \multirow{1}{*}{\textit{Three$\uparrow$}}  &\multirow{1}{*}{\textit{Four$\uparrow$}}  \\ \hline
  \texttt{\#questions} & 220 & 594 & 182 & 503 & 493 & 996 \\ \hdashline      
\textit{random guess accuracy} & 49.55\%  & 28.59\% & 28.31\% & 34.10\% & 29.17\% & 31.47\% \\ \hdashline
 (\texttt{Sep/2023}) LLaVA-v1.5-13B (\textit{length: 2048$\to$2560}) & 47.51\% & 40.74\% & 52.49\% &46.81\% & 41.90\%* & 44.38\% \\
 (\texttt{Oct/2023}) BakLLava (\textit{length: 2048$\to$2560}) & 68.35\%& 35.01\% & 52.78\% & 48.51\% & 42.54\%* & 45.56\% \\
 (\texttt{Nov/2023}) mPLUG-Owl2 \textit{(baseline of \ours)} & 62.25\% & 35.70\% & 53.71\% &44.19\% & 45.42\% & 44.80\% \\
 (\texttt{Feb/2024}) InternLM-XComposer2-VL (\textit{length: 4096$\to$5120}) & 62.95\% & 47.29\% & 52.02\% & 55.70\% & 46.51\%* & 51.76\%  \\ \hdashline
        {Qwen-VL-Max} \textit{(Proprietary)}  &  62.33\% & 70.00\% & \gb{81.54\%} &  72.35\% & 68.79\% & 70.55\% \\
        {Gemini-Pro} \textit{(Proprietary)} & 75.00\% &  67.37\% &  66.92\% & 68.71\% & 70.87\% & 69.79\% \\ 
         {GPT-4V} \textit{(Proprietary, teacher of \ours)} & \gb{80.32\%} & \textbf{77.28\%} & 78.82\% & \textbf{80.32\%} & \textbf{77.28\%} & \textbf{78.82\%} \\ \hline
         \textit{Non-expert \textit{Human}} & 82.27\% & 78.15\% & 74.31\% & 77.18\% & 79.55\% &78.35\%\\ \hline
         \textit{without Multi-image Comparative Data}  & 62.72\% & 37.54\% & 53.30\% & 45.33\% & 46.65\% & 45.98\% \\ \hdashline
         \ours~(Ours) & \textbf{79.55\%} & \gb{85.35\%} & \textbf{81.32\%} & \gb{84.69\%} & \gb{81.94\%} & \gb{83.33\%} \\ \hline
   \end{tabular}}
    \label{tab:micmcq}
    \vspace{-1.8em}
\end{table}

As is shown in Tab.~\ref{tab:micmcq}, \ours~provides very competitive accuracy on open-question quality comparison among three/four images, \textbf{5.7\%} better than GPT-4V (\textit{best existing}) and \textbf{6.4\%} more accurate than non-expert human; open-source LMMs even struggle to obtain 50\% accuracy on this setting. It is also noteworthy that LLaVA series and XComposer2-VL's original context lengths are \textcolor{red}{\textit{not enough for four images}} as they have not reduced visual token numbers, so we have to extend their context windows to evaluate them for \mic. Consequentially, all these models have experienced notably worse accuracy on groups of four (\textit{on extended context length}) than groups of three (\textit{within original context length}), as noted in * in Tab.~\ref{tab:micmcq}. This degradation highlights the importance to \textit{reduce visual tokens} (Sec.~\ref{sec:mtd}) to adapt to multi-image scenarios.


\subsection{Ablation Studies}

\noindent \textbf{Ablation on Training Data.} As the proposed dataset is composed of three parts, we discuss the effects on different subsets of data on the six evaluation scenarios in Tab.~\ref{tab:abldata}. Through the results, we have arrived at several important conclusions: \textbf{{A)}} Even with all information from Q-Instruct-200K, the incorporation of \textbf{Merge2Compare} ({variant (2)}) still notably enhances the capability across various settings; \textbf{{B)}} Only involving the \textbf{Teach2Compare} (\ie GPT-4V labels, {variant (5)}) cannot outperform its teacher GPT-4V; \textbf{\textit{C)}} instead, the superiority towards GPT-4V on question-answering benchmarks is benefited from the co-instruction on \textit{hetero-sourced} training subsets (\textit{variants (6) and (7)}).

\begin{table}[]
\centering
    \renewcommand\arraystretch{1.14}
    \renewcommand\tabcolsep{9pt}
    \caption{\textbf{Ablation Study} on data. \textit{Abbreviation:} \underline{Merge}: \textbf{Merge2Compare}; \underline{{Teach}$^\textit{G}$}: \textbf{Teach2Compare}-\textit{general}; \underline{{Teach}$^\textit{QA}$}: \textbf{Teach2Compare}-\textit{Q\&A}; \underline{QIn}: Q-Instruct-200K.}

    \centering
    \vspace{-10pt}
    \resizebox{\linewidth}{!}{\begin{tabular}{l|cccc|c|c|c|c|c}
    \hline
        \multicolumn{5}{c|}{\textbf{Training Subset}} & \multicolumn{5}{c}{\textbf{Evaluation Scenario}} \\ \hline
        Variant No. & QIn &  Merge & Teach$^\textit{G}$ & Teach$^\textit{QA}$ & Q-Bench$^\texttt{PAIR}$-A1 & Q-Bench$^\texttt{PAIR}$-A2 & Q-Bench$^\texttt{SINGLE}$-A1 & 2AFC-LMM & \mic \\ \hline
        \multicolumn{5}{c|}{\textit{Reference Results of GPT-4V}} & \textcolor{gray}{78.07} & \textcolor{gray}{4.89} & \textcolor{gray}{74.10} & \textcolor{gray}{0.721} & \textcolor{gray}{78.82}  \\ \hdashline
        ({1}) & \cmark &  & & & 53.15 & 3.69 & 73.38 &  0.449 & 45.98 \\
        ({2}) &  \cmark & \cmark & & & 68.67 & 4.67 &  75.12 & 0.701 & 60.34  \\
         ({3}) & \cmark &  &  \cmark &  & 65.44 & 4.64 & 74.38 & 0.647 & 64.82   \\ 
        ({4}) & \cmark & \cmark & \cmark &  & 69.87 & \textbf{4.82} & 76.52 & 0.749 & 66.37 \\ \hdashline
        ({5}) & \cmark & & \cmark & \cmark & 78.28 & 4.65 & 75.72 & 0.676 & 76.41 \\ 
        ({6}) & \cmark &  \cmark & & \cmark & 80.08 & 4.68 & 75.65 & 0.728 & 81.82 \\ 
        \hline
        
        (7, \textit{full}) & \cmark & \cmark & \cmark & \cmark & \textbf{80.18} & \textbf{4.82} & \textbf{77.31} & \textbf{0.754}  & \textbf{83.33}   \\ \hline
    \end{tabular}}
    \label{tab:abldata}
    \vspace{3pt}
    \renewcommand\arraystretch{1.1}
    \renewcommand\tabcolsep{6.6pt}
    \caption{\textbf{Ablation Study} on the text-image interleaved format for the  \ours.}

    \centering
    \vspace{-10pt}
    \resizebox{\linewidth}{!}{\begin{tabular}{l|c|c|c|c|c}
    \hline
        \multicolumn{1}{c|}{\textbf{Format}} & Q-Bench$^\texttt{PAIR}$-A1 & Q-Bench$^\texttt{PAIR}$-A2 & Q-Bench$^\texttt{SINGLE}$-A1 & 2AFC-LMM & \mic \\ \hline
        {\tt <img$_\texttt{0}$>(<img$_\texttt{1}$> ...)} {(\textit{baseline}, popular strategy~\cite{iblip,improvedllava,minigpt4})} & 76.37 & 4.73 & 74.11 & 0.729 & 78.92 \\ \hdashline
         {\tt <img\_st><img$_\texttt{0}$><img\_end> (<img\_st><img$_\texttt{1}$><img\_end>$\dots$)} & 76.77 & 4.79 & 73.85 & 0.736 & 80.12  \\ 
        {\tt The input image: <img$_\texttt{0}$> (The input image: <img$_\texttt{1}$>$\dots$)} & 78.28 & 4.80 & 75.12 & 0.749 & 82.22   \\ \hline
       \textcolor{deepgreen}{{\tt The first image: <img$_\texttt{0}$> (The second image: <img$_\texttt{1}$>$\dots$)}} & \textbf{80.18} & \textbf{4.82} & \textbf{77.31} & \textbf{0.754}  & \textbf{83.33}   \\ \hline
    \end{tabular}}
    \label{tab:ablinterleave}
    \vspace{-1.2em}
\end{table}

\begin{figure}
    \centering
    \includegraphics[width=0.93\linewidth]{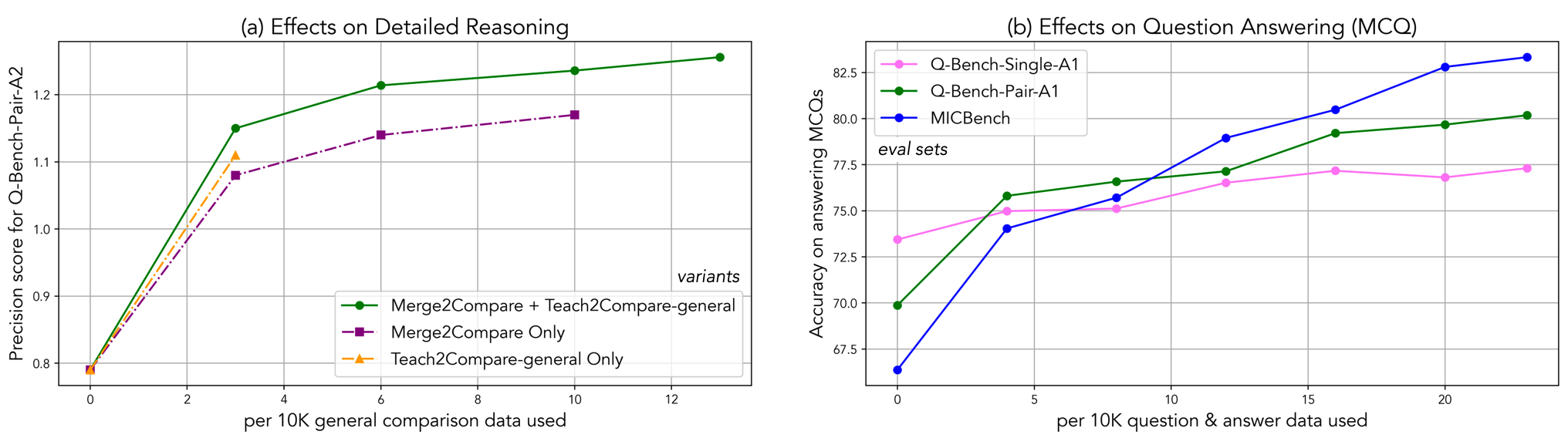}
    \vspace{-1em}
    \caption{\textbf{Effects of Data Scaling} for (a) \textit{detailed reasoning} and (b) \textit{question answering}.}
    \label{fig:datascaleup}
    \vspace{-2.2em}
\end{figure}

Besides the composition of three subsets, we also explore the effects of data scale: in Fig.~\ref{fig:datascaleup}(a), we confirm that more general comparison data contributes to an increase in detailed reasoning capabilities, and mixed data has a better effect than homogenous data at the same scale; in Fig.~\ref{fig:datascaleup}(b), we also validate that scaling up the Q\&A subset helps to improve multiple MCQ metrics.


\noindent \textbf{Ablation on Interleaved Format.} As shown in Tab.~\ref{tab:ablinterleave}, the proposed text-image interleaved format has proved non-negligible advantages than the baseline, as well as other variants \textit{without explicitly noting} image orders. The results suggest the rationale of the format on multi-image comparative settings.

\section{Conclusion}

In this work, we investigate the \textit{open-ended visual quality comparison} problem, with the aim of a model that provides answers and \textcolor{deepgreen}{\textit{detailed reasonings}} on \textcolor{deepgreen}{\textit{open-range questions}} that compares quality among multiple images. To achieve this, we collect the first instruction-tuning dataset to fine-tune large multi-modality models (LMMs) for comparison, the Co-Instruct-562K, with two subsets from human annotations on single images (\textit{merged by LLMs}), and GPT-4V responses.  With the dataset, we propose the \ours, which not only outperforms all existing LMMs (including \textit{its teacher}, GPT-4V) on visual quality comparison, but also marks the first LMM with the capability to surpass human accuracy on related settings. We further construct the \mic, the first benchmark that evaluates multi-image quality comparison for LMMs on three and four images. We expect our work to motivate future studies on visual quality comparison.

\clearpage  

%
%
\bibliographystyle{splncs04}
\bibliography{main}
\end{document}